\documentclass[journal]{IEEEtran}
\IEEEoverridecommandlockouts
\usepackage{cite}
\usepackage{amsmath,amssymb,amsfonts}
\DeclareMathOperator*{\argmax}{arg\,max}

\usepackage{algorithmic}
\usepackage{graphicx}
\usepackage{textcomp}
\usepackage{xcolor}
\usepackage{url}
\usepackage{hyperref}
\def\BibTeX{{\rm B\kern-.05em{\sc i\kern-.025em b}\kern-.08em
    T\kern-.1667em\lower.7ex\hbox{E}\kern-.125emX}}

\usepackage{subcaption}
\usepackage{color}

\usepackage[linesnumbered,ruled,vlined]{algorithm2e}
\newlength{\twosubht}
\newsavebox{\twosubbox}

\title{
REAL-X --- Robot open-Ended Autonomous Learning Architectures: Achieving Truly End-to-End Sensorimotor Autonomous Learning Systems
\thanks{This research has received funding from the European Union Horizon 2020 Research and Innovation Programme under Grant Agreement 
No 713010, 
Project 
\textit{Goal-Robots -- Goal-based Open-ended Autonomous Learning Robots}, \url{http://www.goal-robots.eu/}.
}
}

\author{

\IEEEauthorblockN{Emilio Cartoni$^+$, Davide Montella$^+$, Jochen Triesch°, Gianluca Baldassarre$^+$}\\

\IEEEauthorblockA{
\textit{$^+$Institute of Cognitive Sciences and Technologies}
\textit{National Research Council, Rome, Italy}\\
\textit{gianluca.baldassarre@istc.cnr.it, davidelmontella@gmail.com, emilio.cartoni@istc.cnr.it}\\ 
\textit{° Frankfurt Institute of Advanced Studies, Frankfurt, Germany}\\
\textit{triesch@fias.uni-frankfurt.de}
}
}
    
\begin{document}

\maketitle

\begin{abstract}
Open-ended learning is a core research field of developmental robotics and AI aiming to build learning machines and robots that can autonomously acquire knowledge and skills incrementally as infants and children.
The first contribution of this work is to study the challenges posed by the previously proposed benchmark `REAL competition' aiming to foster the development of truly open-ended learning robot architectures.
The competition involves a simulated camera-arm robot that:
(a) in a first `intrinsic phase’ acquires sensorimotor competence by autonomously interacting with objects;
(b) in a second `extrinsic phase’ is tested with tasks unknown in the intrinsic phase to measure the quality of knowledge previously acquired.
This benchmark requires the solution of multiple challenges usually tackled in isolation, in particular exploration, sparse-rewards, object learning, generalisation, task/goal self-generation, and autonomous skill learning.
As a second contribution, we present a set of `REAL-X' robot architectures that are able to solve different versions of the benchmark, where we progressively release initial simplifications.
The architectures are based on a planning approach that dynamically increases abstraction, and intrinsic motivations to foster exploration.
REAL-X achieves a good performance level in very demanding conditions. We argue that the REAL benchmark represents a valuable tool for studying open-ended learning in its hardest form.
\end{abstract}


\section{Introduction}
\label{Sec:Introduction}

%Background and impact.
The satisfaction of biological and social needs represents a major drive for animal learning.
Beyond these drives, learning in humans and other animals with more sophisticated cognition is also guided by drives such as \textit{curiosity} and \textit{intrinsic motivations}, leading them to acquire knowledge and skills that can be later used to accomplish biological and social needs \cite{Berlyne1960,Baldassarre2011WhatAreIntrinsicMotivationsABiologicalPerspective,GottliebOudeyerLopesBaranes2013Informationseekingcuriosityandattentioncomputationalandneuralmechanisms}.
In the last two decades, several algorithms have been proposed to reproduce such processes in artificial intelligent machines and robots so as to allow them to undergo \textit{open-ended learning}, that is, to autonomously acquire knowledge and skills without relying on pre-wired reward functions, tasks, or goals \cite{Baldassarre2013Book,DoncieuxFilliatDiazRodriguezHospedalesDuroConinxRoijersGirardPerrinSigaud2018OpenEndedLearningaConceptualFrameworkBasedonRepresentationalRedescription}. 
The autonomous acquisition of knowledge and skills in an open-ended fashion has been studied under different headings.
The field of \textit{developmental robotics} \cite{Lungarella2003developmental,Cangelosi2015developmental} has developed algorithms for autonomous learning based on \textit{intrinsic motivations} 
(e.g., \cite{Barto2004,SchembriMirolliBaldassarre2007EvolvingChildhoodsLengthandLearningParametersinanIntrinsicallyMotivatedReinforcementLearningRobot,Oudeyer2007intrinsic,Schmidhuber2010,zhao2012unified,Santucci2014,Tanneberg2019,eckmann2020active,BellasDuroFainaSouto2010MultilevelDarwinistBrainMDBArtificialEvolutioninaCognitiveArchitectureforRealRobots}).
More recently, also machine learning and robotics have started to propose systems to face the challenges of open-ended learning (see Section
\ref{Sec:RelatedModels}), in particular based on reinforcement learning algorithms \cite{Sutton1998,Barto2003}.

An important trend\footnote{Whole projects have been dedicated to this, see for example the EU project \textit{GOAL-Robots -- Goal-based Open-ended Autonomous Learning}: \url{www.goal-robots.eu}} within both fields has been the use of intrinsic motivations not to directly learn skills but rather to guide the self-generation or discovery of \textit{goals}, namely internal representations of the world that might drive the agent's actions to realize these world states \cite{Santucci2013iccm,Santucci2014icdl,Forestier2017,NairPongDalalBahlLinLevine2018VisualReinforcementLearningwithImaginedGoals}. 
The idea is that the autonomous setting of goals can support open-ended leaning as it allows the autonomous generation of tasks which in turn can drive the acquisition of the skills directed to pursue the goals.
An increasing number of works thus focuses on the development of agents able to autonomously form new goals and
learn the associated skills for realizing these goals
\cite{Held2017,Meeden2017,NairPongDalalBahlLinLevine2018VisualReinforcementLearningwithImaginedGoals,Rolf2014,Santucci2016,Seepanomwan2017,KimConinxDoncieux2021FromExplorationtoControlLearningObjectManipulationSkillsthroughNoveltySearchandLocalAdaptation}. New goals may be defined based on the saliency of world states 
\cite{Barto2004}, the change of states
\cite{Santucci2016,Sperati2017}, eigenoptions 
\cite{Machado2017}, density models 
\cite{Bellemare2016}, entropy 
\cite{Eysenbach2018DiversityIA}, or variational inference \cite{DBLP:journals/corr/abs-1807-10299}.
While these are important developments, at present we still do not have systems able to undergo a truly open-ended learning process.

One way to promote the development of such systems is the proposal of benchmarks and competitions that facilitate the comparison of alternative approaches and models.  
Some existing competitions face issues relevant for robotic open-ended learning, but all seems to lack some key elements.
For example, the \textit{AutoML for Lifelong Machine Learning}\footnote{\url{http://automl.chalearn.org/life-long-learning}} competition focuses on the acquisition of an increasing amount of input-output data furnished externally, but it does not involve embodied systems interacting with a physical world to actively generate new experiences.
\textit{Animal-AI Olympics}\footnote{\url{http://animalaiolympics.com}} is focused on simulated animal-like robots, but these are tested with tasks defined through specific reward functions.
The \textit{ICDL MODELbot Challenge}\footnote{\url{https://icdl-epirob2019.org/modelbot-challenge}}
is focused on autonomous developmental processes, but it is not conceived as a benchmark allowing quantitative comparisons among competing approaches.

For these reasons, we have recently proposed a competition, now in its third edition
(\textit{REAL 2021 -- Robot open-Ended Autonomous Learning}\footnote{\url{https://www.aicrowd.com/challenges/real-robots-2020}}),
proposing a benchmark encompassing the major challenges of robotic open-ended learning \cite{CartoniMannellaSantucciTrieschRueckertBaldassarre2020REAL2019RobotOpenEndedAutonomousLearningCompetition}).
The core structure of the competition is based on two phases \cite{CartoniMannellaSantucciTrieschRueckertBaldassarre2020REAL2019RobotOpenEndedAutonomousLearningCompetition,BaldassarreLordGranatoSantucci2019AnEmbodiedAgentLearningAffordanceswithIntrinsicMotivationsandSolvingExtrinsicTaskswithAttentionandOneStepPlanning}: an intrinsic phase of learning and an extrinsic phase of testing.
In the first \textit{intrinsic phase}, involving a very long period of learning during which no guidance is available in the form of rewards, tasks, goals, etc., the robot should autonomously interact with the environment to acquire as much knowledge and skills as possible to best solve any future tasks in the succeeding extrinsic phase.
During the \textit{extrinsic phase}, the knowledge that the robot acquired during the intrinsic phase is evaluated by asking it to solve a certain number of tasks (goals to be achieved) that are unknown during the intrinsic phase.
Since during the intrinsic phase the robot is not given any guidance, the only element that it can exploit to learn knowledge useful to solve the tasks of the second phase is that the environment and object properties are the same in the two phases.  
The competition features two difficulty ``Levels'' in which the participants can take part:
(1) an `Easy Level', allowing the use of some simplifications such as the availability of the position objects on the working space and the possibility of using parameterized motor primitives to control the robot arm;
(2) a `Difficult Level', where the robot has to autonomously acquire end-to-end solutions directly mapping the image pixels and arm joint angles to the desired arm motor commands.   
The competition structure and the general setting is reviewed in more detail in Section~\ref{Sec:ScenarioBenchmark}. 

The benchmark poses extremely hard challenges, as revealed by the previous editions of the competition \cite{CartoniMannellaSantucciTrieschRueckertBaldassarre2020REAL2019RobotOpenEndedAutonomousLearningCompetition}.
The first challenge is that the robot perceives only pixels, joint angles, and touch-sensor readings and so needs to autonomously understand from scratch the very concept of object and which specific objects populate the environment.
Second, the robot does not have any information on how to control its several degrees of freedom to produce `relevant' effects on the environment, for example to touch the objects, let alone to control them in any meaningful way.
Third, and and even worse, the robot has to face a formidable bootstrapping chicken-egg problem as to make sense of objects it should be able to physically interact with them; but at the same time to be able to interact with those objects it should first know them. This circular interdependence of the learning processes of sensory and motor abilities makes it very difficult to acquire them at the same time.

The REAL benchmark thus uncovers the challenges posed by open-ended-learning in the purest and hardest form.   
This is important for both the study of development in children, and for the design of autonomous robots.

Regarding the study of development, the challenges posed by the REAL benchmark reflect in a `pure form' (i.e., neglecting innate behaviours and experience acquired in the womb) the ``great blooming, buzzing confusion'' \cite{James1980PrinciplesofPsychology} that newborn babies have to face at birth when they are suddenly immersed in a flood of unstructured information from the senses, and have to control the many degrees of freedom of their bodies.
This problem has been studied several times within the Developmental Robotics community, but very often it is simplified by furnishing the robots with some hardwired knowledge.
Instead, the Difficult Level of the REAL challenge does not give the robot any pre-wired knowledge and any guidance for learning. 
In this respect, one might argue that the challenge is even too hard as the development of babies and children can rely on important forms of social scaffolding such as imitation, joint attention, social rewards, and teaching. 
In this respect, the REAL benchmark purposefully leaves out such forms of scaffolding for the sake of focusing on how much autonomous individual learning mechanisms can accomplish without using any form of external information and guidance. 

In addition, the benchmark is also important to develop autonomous robots able to undergo open-ended learning to face real-life non-engineered human scenarios as those that might be encountered in homes or workplaces.
These scenarios are particularly demanding as they pose challenges that cannot be foreseen at design time.
For example, a robot-assistant able to help to tidy up a kitchen will need sufficient versatility to adapt to different kitchen layouts, kitchen objects etc.
Thus, instead of coming out of the factory with pre-defined skills it should leverage intrinsic motivations and self-generated goals/tasks to acquire in the working environment a repertoire of skills that will be useful to accomplish human assigned tasks \cite{Seepanomwan2017}. 
The application value of these type of algorithms cannot be overestimated.
Indeed, a robot system able to undergo truly open-ended learning could in principle be employed in any different environments, with different robotic platforms, and to pursue any goal relevant for the users.
A first contribution of this work is indeed to contribute to clarify the actual nature and difficulty of the challenges posed by the REAL open-ended learning benchmark from a robotic point of view.

The literature has proposed various models recently, each having some features that might be relevant to face the REAL benchmark.
These models are briefly introduced here and then considered in more detail in Section \ref{Sec:RelatedModels}.

Some models \cite{Pathak2017, Burda2018, Pathak2019, Sekar2020} focus on exploration using a reward based on prediction errors.
However, all these models use reward functions to guide the agent to execute a task and they do not learn a goal-conditioned policy that can be readily used to reach goal states: to face the REAL benchmark, they would need an added component that translates the goal images into suitable reward functions. 
The models presented in \cite{Nair2018,Pong2019} also focused on exploration, in this case using a variational auto-encoder to generate target states and learn the skills to accomplish them. These models can potentially be applied to the REAL benchmark but this is prevented by the limitations discussed below.
Another work \cite{Florensa2017} progressively trains an agent by using a network to predict states with `intermediate difficulty' to be reached, thus focusing on `competence' instead of `novelty' \cite{SantucciBaldassarreMirolli2013Whichisthebestintrinsicmotivationsignalforlearningmultipleskills}.
The model is not compatible with the REAL benchmark, as it requires the environment to be reset periodically during the training.
Other relevant works \cite{Ding2020,Laskin2019,Yu2019,Wang2019} are based on image-based planning and use a controller to execute plans formulated on the basis of latent state representations. 

Among these models, \cite{Ding2020} seems not applicable to the REAL scenario because it lacks the mechanisms needed to support autonomous learning during the intrinsic phase since it relies on an external reward objective.
Instead, the works of \cite{Laskin2019,Yu2019,Wang2019}, and also of  \cite{Nair2018, Pong2019} mentioned above, seem to be applicable to the REAL challenge.
This however would need to adapt their code (available for \cite{Laskin2019,Yu2019,Nair2018, Pong2019}) to the benchmark scenario, a task that might be carried out in future work if it will be possible to solve the implementation issues presented in the Section \ref{Sec:RelatedModels}.

A second contribution of this work is to propose a set of architectures, called \textit{REAL-X}, that encompass a number of solutions to face the different challenges posed by the REAL benchmark.
These represent the first models achieving a score beyond chance level in REAL, in particular by learning to move objects closer to their goal positions.
The `X' in the name refers to the fact that the architecture has been developed in different versions built in an incremental fashion to face increasingly difficult conditions of the REAL scenario.
In particular, we first faced simplified versions of the REAL challenge with a basic version of the architecture, and then we progressively removed the simplifications and faced them by endowing the architecture with enhanced components.
The simplifications considered included: 
(a) the use of information on the positions of objects rather than raw-pixel images; 
(b) the use of a parameterised macro-action rather than joint angles; 
(c) the use of only one object rather than two or more.

The REAL-X architecture has a modular design to facilitate incremental development and is in particular formed by three components:
(1) \textit{Abstractor}: this component performs the abstraction of sensory inputs to learn relevant environment variables, for example related to the position and identity of objects;
(2) \textit{Explorer}: this component generates the motor experience supporting the learning of goals and actions during the intrinsic phase;
(3) \textit{Planner}: this component formulates and executes action plans to accomplish the extrinsic goals.
The tests of the different versions of REAL-X show that they exhibit coherent behaviour and achieve a performance well above chance level with the simplifications, and also when these simplifications are progressively removed.
In the following we present a thorough investigation of the challenges posed by REAL benchmark and possible solutions relying on the REAL-X architecture; in future work we plan to use the REAL benchmark and the architecture to further develop these solutions and systematically compare and integrate them with the models and mechanisms reviewed above and in Section \ref{Sec:RelatedModels}. 

The rest of the paper is organised as follows.
Section~\ref{Sec:Methods} first presents more in detail the open-ended learning REAL benchmark and its objective function, and then presents the baseline REAL-X architecture and its different variants created to face increasingly complex version of the scenario.
Section~\ref{Sec:Results} compares the performance of the different REAL-X architectures and analyses some aspects of their internal functioning.
Section~\ref{Sec:RelatedModels} reviews in more depth relevant previous models on open-ended learning.
Finally, Section~\ref{Sec:Conclusions} draws the conclusions.

%***************************************************************************
\section{Methods}
\label{Sec:Methods}

\subsection{Scenario of the open-ended learning benchmark}
\label{Sec:ScenarioBenchmark}

\begin{figure}[htp]
\sbox\twosubbox{%
  \resizebox{\dimexpr1.2\textwidth-1em}{!}{%
    \includegraphics[height=3cm]{example-image-a}%
    \includegraphics[height=3cm]{example-image-16x9}%
  }%
}
\setlength{\twosubht}{\ht\twosubbox}
% typeset
\centering
\subcaptionbox{REAL Environment\label{Figure:REAL2021}}{%
  \includegraphics[height=\twosubht]{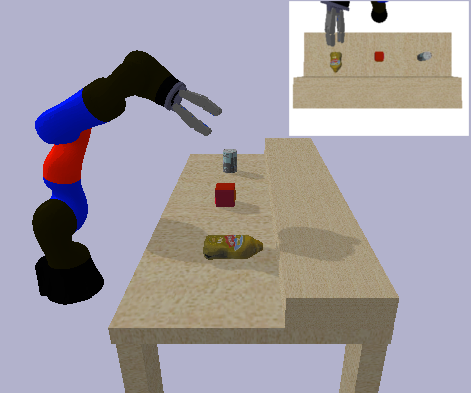}%
}\quad
\subcaptionbox{Examples of extrinsic goals\label{Figure:REAL-goals}}{%
  \includegraphics[height=\twosubht]{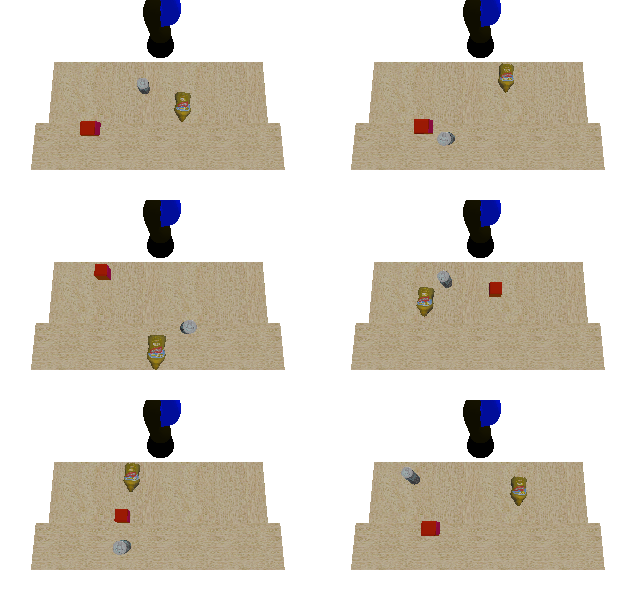}%
}
\caption{
(a) The environment encompassing the robot, the table with the shelf, and the three objects. The inset shows the environment as seen from the robot top-view camera.
(b) Three examples of the 50 tasks used in the extrinsic phase: each row shows the initial object configurations (left) and the final ones, that is, the \textit{extrinsic goals} (right), for the different tasks.}
\end{figure}

The competition scenario is inspired by an assistant-robot scenario where the robot should help to tidy up a kitchen \textit{(`kitchen scenario')}.
Specifically, the architectures participating in the competition have to control a camera-arm-gripper robot manipulating some objects on a table simulated with a physics engine (see Figure~\ref{Figure:REAL2021}).
The robot is a 7-DoF Kuka arm coupled with a 2 DoF gripper.
It stands in front of a table, which has a shelf and $1$ to $3$ objects: a cube, a tomato can, and a mustard bottle.
The perceptual space comprises the images from a fixed top-view camera, the proprioception of joint angles, and the gripper touch sensors; in a simplified version of the benchmark the robot directly perceives the positions of the objects.
In the REAL-X architectures discussed below, the touch sensors will not be used.
The action space is the arm-gripper joint space;  however, in a simplified version of the benchmark the robot can be controlled based on a parameterised macro-action.
The environment also allows control of the arm in the Cartesian space combined with control of the gripper in the joint space.
However, in the REAL-X architectures discussed below we used only the macro-action and whole-body joint control.

%Core of benchmark: intrinsic and extrinsic phases
The kitchen scenario is inspired by a possible future real world use case for open-ended learning robots. 
In this regard, one of the main objectives of the competition is to translate this kind of scenario into \textit{a robotic benchmark for objectively measuring the capacity of robots for autonomous open-ended learning}.
To this purpose, a main feature of the benchmark is its organisation into two phases \cite{BaldassarreLordGranatoSantucci2019AnEmbodiedAgentLearningAffordanceswithIntrinsicMotivationsandSolvingExtrinsicTaskswithAttentionandOneStepPlanning,CartoniMannellaSantucciTrieschRueckertBaldassarre2020REAL2019RobotOpenEndedAutonomousLearningCompetition}.
In a first \textit{intrinsic phase}, the robot \textit{autonomously} interacts with the environment for a long time during which it should acquire as much knowledge and skills as possible to best solve the tasks in a second `extrinsic phase'.
Here, the intrinsic phase lasts $15,000,000$ simulation steps each lasting $5$ ms, amounting to $20.83$ hours of simulated time.
During the \textit{extrinsic phase}, the acquired knowledge and skills are evaluated with tasks (here `goals', intended as desired world states) unknown by the robot during the intrinsic phase.
The extrinsic goals are drawn from the following classes of possible object configurations:
(a) \textit{2D goals}: goals defined in terms of the configuration of 1 to 3 objects on the table plane, never close to each other, and with a fixed orientation;
(b) \textit{2.5D goals}: goals defined in terms of the configuration of 1 to 3 objects set on the table plane and on the shelf, never close to each other, and with a fixed orientation;
(c) \textit{3D goals}: goals defined in terms of 1 to 3 objects set on the table plane and on the shelf, with any orientation and no minimum distance.
Each goal involves a different starting configuration, which adheres to the same criteria as for the goal itself.
Figure~\ref{Figure:REAL-goals} shows some examples of extrinsic goals.
Crucially, in the extrinsic test the robot can still learn, but this is of little help given the limited time available to solve each task, so \textit{the performance in the extrinsic phase can be considered an objective measure of the system's capacity to autonomously acquire knowledge during the intrinsic phase}. 
Importantly, during the intrinsic phase the robot is not given any knowledge: no tasks, reward functions, pre-trained networks for object recognition, world models, abilities, or motor skills.
The robot learning processes used in the intrinsic phase should hence be guided by algorithms supporting autonomous learning, for example intrinsic motivations and mechanisms for the self-generation of tasks or goals.

%Objective function
\subsection{Objective function of the open-ended learning benchmark}
The overall objective of the robot participating in the competition is to find, \textit{during the intrinsic phase}, the parameter vector $\theta^*$ that maximizes the expected reward collected \textit{during the extrinsic phase}:
\begin{equation}
  \theta^* = \argmax_\theta E_{g\sim\tau(g)} \left(E_{\pi(a|s,g,\theta)}R(g)\right) 
\end{equation}
where
$R(g)$ are the total rewards obtained for goal $g$ used in the extrinsic phase to evaluate the system,
$g\sim\tau(g)$ is the distribution of possible tasks that can be posed in the environment,
$\pi(a|s,g,\theta)$ is the control policy, dependent on parameters $\theta$, that the robot uses to select actions $a$ in response to state $s$ and the currently pursued goal $g$. 
\textit{The crucial feature of the benchmark is that the parameters} $\theta$ \textit{must be learned during the intrinsic phase but are tested with goals} $g$ \textit{during the extrinsic phase, and these goals are unknown during the intrinsic phase}. 
The latter condition implies that the knowledge that the agent can pass from the intrinsic to the extrinsic phase must rely on the \textit{physics of the environment and the objects that remain the same in the two phases}.
The optimal policy $\pi(a|s,g,\theta^*)$ also depends on the time that the robot has to solve each goal $g$ (see \cite{VermeSilvaBaldassarre2020OptimalOptionsforMultiTaskReinforcementLearningunderTimeConstraints}), but for simplicity this issue is not considered here.

\subsection{Metrics}
During the extrinsic phase the system is asked to solve $50$ tasks (Figure~\ref{Figure:REAL-goals}).
For each task:
(1) the robot is shown a certain configuration of the $3$ (or $2$ or $1$) objects in the environment (`overall goal') in which they are placed anywhere on the table plane or on the shelf: the objects can have any initial/final orientation and may touch/overlap;
(2) the objects are then set in a different position and orientation in the environment;
(3) the robot is given $10,000$ simulation steps ($50$ s) to bring the object(s) to the overall goal configuration. 

The extrinsic-phase performance for an overall goal $g$ is scored with the metric $M_g$:
\begin{equation}
 M_g = \sum_{o=1}^n \left[e^{-c||\textbf{p}^*_o - \textbf{p}_o||} \right]\\
\end{equation}
where $n$ is the number of objects (1, 2, or 3), $\textbf{p}_o^*$ is the $(x, y, z)$ position vector of the center of mass of object $o$ in the overall goal,
$\textbf{p}_o$ is the position of the object at the end of the task after the robot attempts to bring it to the goal position,
$c$ is a constant ensuring that this part of the score will be $0.25$ if the distance to the specific object goal position is $10$ cm.
Note that the metric ranges in $(0,1]$ for each object, is equal to $1.0$ if the object is exactly at the goal position, and decays exponentially with increasing distance from it.
Placing all $3$ objects exactly in the overall goal configuration yields a maximum score of $3.0$.
The total $Score$ $M$ is the average of the scores across the  $G$ ($G=50$) goals:
\begin{equation}
 M = \frac{1}{G} \sum_{g=1}^G M_g\\
\end{equation}
For simplicity the score does not consider object orientation, but keeping track of orientation allows a finer object control and hence facilitates a higher final score, so it is implicitly rewarded.

\subsection{Simplifications}
We have designed different REAL-X architectures to face different versions of the described benchmark.
These versions feature different simplifications that were progressively released to move towards the full hardest challenge.
In particular, the simplifications were implemented along three dimensions: perception, motor action, and number of objects.
A fourth dimension involves the absence of environment reset at the end of `trials/rollouts', which is commonly used in reinforcement learning tasks but here is avoided.
These simplifications are now considered more in detail.

\paragraph{Perception} In a first simplification the robot was given the $x,y,z$ positions of the objects.
When this simplification was removed, the robot instead received a raw camera image of 320x240 RGB pixels.
This is a challenging condition as it introduces a difficult `chicken and egg' problem as to acquire information on objects the robot needs to act on them, but to learn to act on the objects the robot needs to know where the objects are in space.

\paragraph{Motor control} In this simplification the robot used a parameterised macro-action to act on the objects. In particular, the macro-action first moved the (closed) end effector close to the working plane, then along a segment trajectory parallel to the plane, and then back to a home position (arm straight up out of the camera sight).
The use of the macro-action greatly facilitated hitting/pushing the objects during exploration.
The parameters of the macro-actions were the two $(x, y)$ and final $(x', y')$ extremes of the movement segment indicating positions of the end actuator on the plane. 
When this macro-action is usable, the robot needs `only' to learn the four parameters of the macro-action based on the objects' state. When this simplification is removed, the robot has to directly set the desired joint angles at each step.
This poses a great challenge as in the initial phase of (random) exploration of the joint space the robot rarely touches the objects.

\paragraph{Number of objects} The number of objects in the environment is another critical element determining the level of difficulty of open-ended learning.
In the simplest case we consider only one object.
In more challenging conditions we consider up to three.
When the position and identity of objects is furnished to the robot, this does not represent a problem as the robot can focus on one object per time and so the situation becomes similar to the case involving only one object.
However, in the most realistic condition where the robot perceives the environment through RGB images having more than one object represents a notable challenge as the robot needs to process the image pixels to make sense of the existence of different objects.
Moreover, the presence of multiple objects can also make the planning and control more difficult as they can interact and interfere with each other.

\paragraph{Environment resets} During the intrinsic phase, when the objects are moved by the robot they are left where they are, without being reset to their initial position periodically.
This element, that at first sight might seem a secondary technical detail, turned out to be one of the most important aspects of the challenge (note how most reinforcement learning tasks make the reset assumption).
Indeed, to develop the different components of REAL-X we often used the simplified condition where we reset the objects although the experiments reported here do not report the results of this condition as it violates a fundamental element of autonomous open-ended learning (e.g., resetting a real environment would require another agent to do so to support the robot learning).
The challenge is that if objects are not reset to the same position/orientation after each action execution, then after each contact of the robot with the object the succeeding initial condition will change.
This implies that initially the robot tends to face a sequence of new situations making it difficult to accumulate knowledge.

\subsection{REAL-X: blueprint architecture}
\label{Sec:Architectures}

\begin{figure}
    \centering
    \includegraphics[width=8cm]{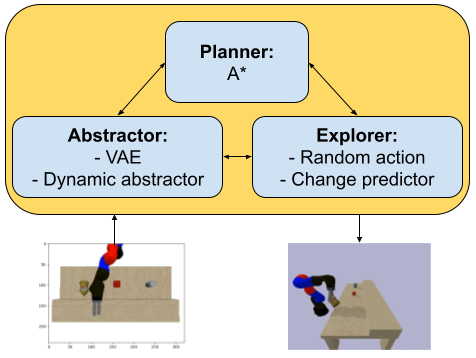}
    \caption{
    General architecture of REAL-X, based on three components: Abstractor, Explorer, and Planner. 
    The boxes indicate the specific solutions used here to implement the components.
    }
    \label{fig:architecture-x}
\end{figure}

In this section we describes the overall REAL-X blueprint architecture, while in the following section we present the variations of the architecture that have been used to face the different challenges emerging when progressively avoiding the simplifications of the REAL scenario.
The REAL-X blueprint architecture is formed by three main components (Figure~\ref{fig:architecture-x}): an explorer module, an abstraction module and a planner module.
These components are able to autonomously learn how to control the environment within the time frame of the intrinsic phase, achieving good control performance within 2000 actions already as shown in the results.
We now consider the functions and main mechanisms implemented by the components, and then we consider their specific enhanced versions forming the different REAL-X  architectures developed to face the increasingly difficult challenges of the REAL benchmark.

\paragraph{Explorer} This component guides motor exploration of the environment during the intrinsic phase in order to maximise the acquisition of motor skills.
Exploration is based on actions each lasting 1000 steps. Each action starts and ends in a `home' position involving the arm and gripper straight upward.
In its basic version, the explorer produces actions by generating random sets of the parameters of an action, either referring to the macro-action or to joint trajectories (explained below). 
On each action the components store the acquired knowledge in the form of a \textit{action triplet} $(s, a, s')$ containing:
(a) the \textit{precondition}, that is, the initial state $s$ of the environment encoded in terms of image or positions of objects;
(b) the \textit{action parameters} $a$ of the performed actions;
(c) the action \textit{outcome}, that is, the state $s'$ that follows the action (image/position of objects).

\begin{figure}
    \centering
    \includegraphics[width=7cm]{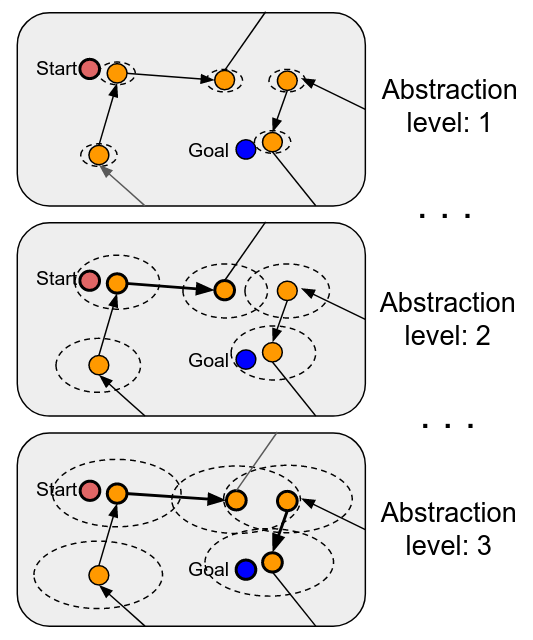}
    \caption{Dynamic abstraction: scheme of functioning. 
    In each of the three rectangles, a different level of abstraction is represented.
    The yellow circles in each rectangle represent the states experienced during the intrinsic phase, the red circle represents the current position of the object seen in the environment, and the blue circle represents the position of the object in the goal image.
    The arrows are the experienced actions that connect the different states.
    The dashed ovals represent the boundary within which a certain state (at the centre of each oval) is considered to be the same as the states within the boundary.
    The oval shape of the dashed circles suggests that at each level of abstraction each state dimension can have different thresholds.
    The bold circles and arrows represent the plan portion that the algorithms managed to find up to a certain level of abstraction: only the last abstraction level allows finding a plan linking the start and goal states.}
    \label{Fig:real-x-scores}
\end{figure}

\paragraph{Abstractor}
We introduced a `Dynamic Abstraction' (DA) mechanism, sketched in Figure~\ref{Fig:real-x-scores}, that defines a number of abstraction levels, starting from the $(s, a, s')$ triplets experienced. The motivating idea is that the planner should be able to reuse the actions of those triplets, even if the current state is not exactly one of starting states ($s$) of those triplets, or even if the goal state is not exactly one of the outcome ($s'$) states.
To define when a certain state $s_i$ can be considered equivalent to an $s$ or $s'$ state of a  triplet, we define a series of increasing thresholds that define different abstraction levels.
These thresholds are constructed by considering the differences caused by the experienced actions, considering in particular the minimum and maximum differences. 
We reasoned that it does not make sense to make a distinction between states that are nearer than the minimum difference that the agent has experienced between each variable, since it has no actions that can move it through states at such a resolution.
Conversely, the highest abstraction should not `merge' states that are farther than the maximum experienced difference since, with such an high threshold, even the action which caused the largest difference would not bring the agent to a different state.
The DA algorithm thus works as follows.

%ARCHITECTURE FUNCTIONING
\begin{algorithm}[h]
\SetNoFillComment
\DontPrintSemicolon
Input: $G$ extrinsic goals,  environment;\\
Output: global\_performance\\
\vspace{0.2cm}
obs\_pre = env.get\_observation()\\
\For{$i \gets 1$ \textbf{to} $I$\tcp*{intrinsic phase steps}}
{
    action $ \gets $ explorer.selectNextAction(obs\_pre)\\
    obs\_post $ \gets $ env.step(action)\\
    transitions.add(obs\_pre, action, obs\_post)\\
    obs\_pre $ \gets $ obs\_post\\
}

abstract\_transitions = VAE(transitions)\\
da = DynamicAbstractor(abstract\_transitions)\\
planner = Planner(da, abstract\_transitions)\\

\ForEach{$s_g \in G$\tcp*{extrinsic phase goals}}
{
  goal\_to\_pursue $ \gets s_g $\\
  obs = env.get\_observation()\\
  \For{$i \gets 1$ \textbf{to} $J$\tcp*{extrinsic goal steps}}
  { 
    action $ \gets $  planner(obs, goal\_to\_pursue)\\
    observation $ \gets $  env.step(action)\\ 
  }
  $s_{final}  \gets $  obs \\
  performance\_array.append(evaluate($s_{final}$,$s_g$))
}
return(global\_performance(performance\_vector))
\caption{Benchmark and architecture functioning}
\label{algo:dynamic_abstractor}
\end{algorithm}

% DYNAMIC ABSTRACTOR
\begin{algorithm}[h]
\SetNoFillComment
\DontPrintSemicolon
Input: $(s, a, s')$ triplets,  L levels of abstractions;\\
Output: abstractions distances matrix\\
\vspace{0.2cm}
\ForEach{$(s, a, s') \in triplets$}{%
    difference\_vector $\gets$ $|s-s'|$
    difference\_matrix.append(difference\_vector)
}
\For{$v \gets 1$ \textbf{to} $V$\tcp*{V variables}}  
{
    sort\_by\_column(difference\_matrix, v)
}
\For{$l \gets 1$ \textbf{to} L\tcp*{L levels of abstraction}}
{
    t $\gets$ 0 \tcp*{T triplets}
    \For{$v \gets 1$ \textbf{to} $V$}{
        abstractions[l, v] $\gets$ difference\_matrix[t, v]\;
    }
    t $\gets$ t + T / (L - 1)
}
\caption{Dynamic abstractor}
\label{algo:dynamic_abstractor}
\end{algorithm}

\begin{algorithm}[h]
\SetNoFillComment
\DontPrintSemicolon
Input: environment, action\_space \\
Output: transitions \\
\vspace{0.2cm}
obs\_pre $ \gets $ env.get\_observation()\\
\For{$i \gets 1$ \textbf{to} $I$\tcp*{intrinsic phase steps}}
{
    \tcp{Bootstrap}
    \If {$i  < 500$} {
        action $ \gets $ action\_space.random\_sample()
   
    } \Else {
        \tcp{Training every 500 actions}
        \If {$i \bmod 500 = 0$} {
            abstract\_transitions = VAE(transitions)\\
            max\_difference = max(abst\_pre - abs\_post)\\
            dataset = $ \varnothing $\\
            \ForEach{pre,post $ \in $ abstract\_transitions}{
                \If {pre - post $ > $ max\_difference * 0.01} {
                    change = 1
                } \Else{
                    change = 0
                }
                dataset.append(pre, action, change)\\
            } 
            network.train(dataset)\\
        }
        \tcp{Action generation}
        \For{$n \gets 1$ \textbf{to} 1000} 
        {
            action $ \gets $ action\_space.random\_sample()\\
            change = network.predict(action)\\
            \If {change = 1} {
                \textbf{break}
            }
        }
    }
    obs\_post $ \gets $ env.step(action)\\
    transitions.add(obs\_pre, action, obs\_post)\\
    obs\_pre $ \gets $ obs\_post
}
\caption{Intrinsic motivation exploration based on the Change predictor.} 
\label{algo:dynamic_abstractor}
\end{algorithm}

The output of the DA is an $L \times V$ matrix, where $L$ is the number of desired abstraction levels (here 200), and $V$ is the number of state variables to consider (e.g. the x-y position of an object; or the latent variables of an auto-encoder used to abstract images, as explained below).
For each level of abstraction, the DA gives $V$ thresholds, one for each variable.
This $L \times V$ matrix is used later by the planner to decide which states are considered equivalent to each other at a given level of the abstraction: two state vectors whose difference is less than the thresholds specified for each variable $v$ at a certain abstraction level $l$ are considered equal, i.e. $s_1 \simeq s_2$ if $ 
|s_1[v] - s_2[v]| < DA[l, v] \quad \forall v \in V$.
To compute the levels of abstraction, the absolute differences $|s - s'|$ of each action triplet are first computed and then ranked, independently for each variable. From these ranked differences, 200 differences, one for each abstraction level, are then selected for each variable, starting from the lowest rank up to the maximum rank, at equal intervals.
The lowest level of abstraction thus involves $V$ thresholds representing the minimum differences found, for each state variable, between all the starting-outcome state couples of the triplets; the maximum level of abstraction uses instead the largest differences found.
This process allows the DA to work in an unsupervised fashion for any domain and requires only a single parameter establishing the granularity of the different abstraction levels (how many levels are desired).
The DA can then be used to pursue a given extrinsic goal based on the planning processes described below.

\paragraph{Planner} This component is used in the extrinsic phase to pursue each single extrinsic goal.
Given an extrinsic goal, the component is able to search sequences of action triplets that lead from the current state of the environment to a desired goal state.
When a plan is found, the first action of the plan is performed and then re-planning is repeated to deal with noisy outcomes.
The planning used was A* with a maximum depth of 10 actions.
The heuristic used for the A* search was the outcome-goal distance based on the L-1 norm.
The planning process works in close integration with the DA mechanism.
The states are initially abstracted at `abstraction level 1':
if a plan is found, its first action is executed, otherwise the planning process tries to find a plan with the succeeding higher level of abstraction, and so on until a plan is found (‘abstraction level 2’, ‘abstraction level 3’ etc.).
If the DA reaches the maximum level of abstraction without finding a plan, the goal is aborted as deemed non reachable with the current action triplets.
By using the DA, planning will be able to generalize the triplet preconditions or triplet outcomes to increasingly different environment states.
A higher abstraction however comes to the cost of less precise actions.

%****************************************************
%Different REAL-X architectures
\subsection{Increasingly sophisticated REAL-X architectures}

We now consider the different REAL-X architectures developed to face increasingly difficult versions of the REAL benchmark.
See Table~\ref{Table:SystemsTests} for a summary of the different systems tested in the increasingly challenging conditions of the benchmark.
These conditions were called $REAL\_X$ for ease of reference (note the character $\_$ in place of $-$). 

\begin{table}[]
    \centering
    \begin{tabular}{c | c c c}
\hline
        \textbf{REAL-X}   & \multicolumn{3}{c}{\textbf{REAL$\_$X benchmark conditions}}\\
    \textbf{models} & \textbf{REAL\_OM} & \textbf{REAL\_IM} & \textbf{REAL\_IJ}\\
\hline
\textbf{REAL-R}   &    $0.021\pm.002$   &   $0.021\pm.002$ &  $0.038\pm.002$ \\
\textbf{REAL-T}   &    $0.212\pm.006$*  &                &   \\
\textbf{REAL-D }  &    $0.216\pm.004$   &    NA          &   \\
\textbf{REAL-LD}  &             &    $0.222\pm.005$       &  $0.139\pm.007$\\
\textbf{REAL-ILD} &             &    $0.234\pm.007$       &  $0.149\pm.006$\\

\hline
\multicolumn{4}{l}{\textbf{Letters in place of the `X' in the model acronym `REAL-X':}}\\
\multicolumn{4}{l}{R - Control system performing random actions in the extrinsic phase}\\
\multicolumn{4}{l}{T - Threshold with fixed value (instead of the Dynamic abstractor)}\\
\multicolumn{4}{l}{D - Dynamic abstractor (to support planning)}\\
\multicolumn{4}{l}{L - Latent variables (of the VAE)}\\
\multicolumn{4}{l}{I - Intrinsic motivation (to guide exploration)}\\
\hline
\multicolumn{4}{l}{\textbf{Letters in place of `X' in the benchmark acronym `REAL$\_$X':}}\\
\multicolumn{4}{l}{O - Objects' position as input}\\
\multicolumn{4}{l}{I - Images as input}\\
\multicolumn{4}{l}{M - Macro-action as control output}\\
\multicolumn{4}{l}{J - Joints as control output}\\
\hline
   \end{tabular}
\caption{Performance of the different architecture configurations, named REAL-X, in different conditions, named as REAL\_X. The upper part of the table reports the performance of the different architectures in the different conditions (mean score and standard error of the mean over 30 repetitions of the test) while the lower part of the table indicates the meaning of the acronyms used in place of the  `X' in the names of the architecture or the conditions.
*: Best result after running with different thresholds. NA: the result is not available as the Planner was not able to return a plan before running out of memory.}
    \label{Table:SystemsTests}
\end{table}

\paragraph{REAL-D: Dynamic abstractor; REAL-T: Threshold; REAL-R: Random} 
The first architecture, REAL-D, was used to face the simplest version of the benchmark where the robot was given the object position and the macro-action.
The components of this architecture were implemented as illustrated above with no specific enhancements.
The innovation of REAL-D is represented by the Dynamic abstractor described above allowing the performance of planning at levels of abstraction autonomously decided by the system.
REAL-D was tested in the REAL\_OM condition involving the information on objects position and the macro-action.
To test the utility of the Dynamic abstractor in this condition we also tested a version of the system, called REAL-T, where we manually set a \textit{fixed} abstraction threshold.
It was not possible to test REAL\_D with images as  the A* Planner was not able to return a plan before running out of memory.
We also compared all systems with REAL-R, a system facing the extrinsic goals by producing random actions, used to have a baseline performance in all conditions.

\paragraph{REAL-LD: Latent variables and Dynamic abstractor}
The REAL-LD architecture was used to face the version of the benchmark where the robot had to use raw images although it could still use the macro-action, the REAL\_IM condition.
The Explorer and Planner components were as in REAL-D.
The Abstractor was instead enhanced to tackle RGB images with an additional abstraction process run at the end of the intrinsic phase.
To this purpose, the precondition and outcome images were first processed with the \textit{OpenCV MOG2} computer vision algorithm filtering everything that does not change much in different images, in our case the background of objects.
The algorithm does this in a fully autonomous way based only on the images collected during the intrinsic phase.
We filtered the background because we assume that the robot is interested in learning about \textit{things that it can change with its actions}, and so the static areas of the image are not interesting for it.
Next, a Variational Autoencoder (VAE; \cite{KingmaWelling2013AutoEncodingVariationalBayes}) is trained with the background-filtered precondition and outcome images of the triplets.
Based on this training, the VAE can extract a compact representation of images encoded by the activation pattern of the VAE bottleneck (latent variables). 
The state latent variable representations are used in place of the images in all planning processes involve in the extrinsic phase.
REAL-LD was an important architecture that allowed to focus on studying the challenges posed by a perception based on raw-pixel images without the complications of the motor aspects.
The model was thus also tested with 2 objects achieving lower performance than with 1 object but higher than random (0.095 vs 0.060 of REAL-R, not further reported).

REAL-LD was also tested in the REAL\_IJ condition involving not only raw-pixel but also the direct control of joints.
To face this condition, the macro-action was substituted with a control method that did not make specific assumptions about how to act in the environment.
In particular, an action was implemented as a sequence of via points and directly used to control the arm-gripper joints (based on the robot's PIDs).
An action was generated as follows. 
The whole action still lasted 1000 steps in total.
First a random number of via points (from 1 to 10) was generated with a uniform distribution.
Then each via point was generated by randomly sampling the joint angles with a uniform distribution.

\paragraph{REAL-ILD: Intrinsic motivation exploration, Latent variables, Dynamic abstractor} 
The REAL-ILD architecture worked as REAL-LD but its Exploration component was enhanced with a mechanism directed to improve the efficiency of exploration during the intrinsic phase based on a new intrinsic motivation mechanism.
This mechanism was in particular used to increase the likelihood that the actions generated for exploration touched the objects.
The mechanism is based on a neural-network predictor that takes as input the current state and a planned action (parameters), and predicts if the action will lead to a \textit{significant change} of the state, for example because an object has been moved.
The predictor is trained every 500,000 steps based on the information acquired that far.
A VAE is also trained every 500,000 steps to feed suitable representations of the images to the predictor,  with the data collected that far.
The predictor is then used before performing each exploration action to select the actions that have a high chance to cause a change (hit the objects).
In particular, the system generates up to 1,000 actions and then performs the first one that is predicted to cause a change or a random one if no such action is found.
%********************
\section{Results}
\label{Sec:Results}

Table~\ref{Table:SystemsTests} reports the scores of all REAL-X architectures tested with their respective versions of the benchmark.
The performance and some aspects of the internal functioning of the architectures are now analysed in detail.
A video of the performance of REAL-LD is available on \url{https://youtu.be/kl26SyGAy_M} (REAL\_IM condition) and on \url{https://youtu.be/nriO73Sftq0} (REAL\_IJ condition).

\paragraph{REAL-D vs. REAL-R and REAL-T: test in the REAL\_OM condition}
The REAL-D architecture, tested in the REAL\_OM condition, achieves a score of 0.216 while a random model REAL-R achieves a performance of 0.021.

The role of the Dynamic abstractor for such result can be seen by comparing such performance to the one of REAL-T.
We tried different values of the threshold, and only the best one led to a performance similar to the one of REAL-D, 0.212.
In particular, we first tried with threshold values close to those most used by REAL-D while planning, that is 1cm and 2cm for respectively the x and y position of the object.
The score, however, was only 0.141. 
We then tested lower values, (0.005, 0.01), which obtained a lower score (0.040), and higher values, (0.02, 0.04) and (0.04, 0.08), which obtained 0.212 and 0.150 respectively.
This shows the capacity of the Dynamic abstractor to automatically find a good level of abstraction compatible with the found actions. 

\begin{figure}
    \centering
    \includegraphics[width=8cm]{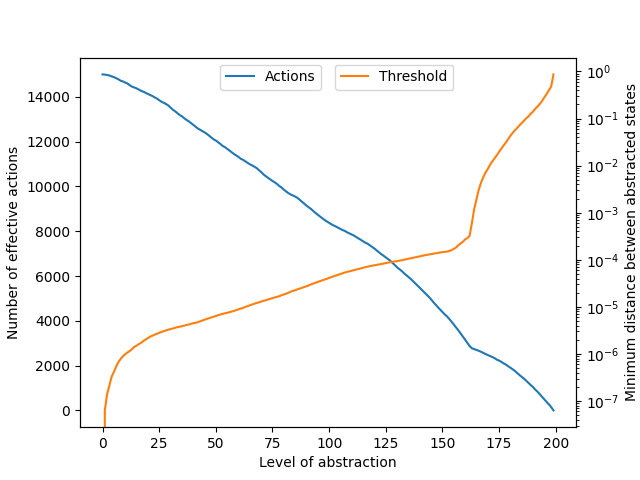}
    \caption{REAL-D: Number of `effective' actions (left y-axis) at different levels of abstraction (x-axis). 
    An action is `effective' if it changes the world from a state to another state that is considered different at the considered level of abstraction.
    The right y-axis measures the minimum distance between two states to be considered different at that abstraction level. 
    } 
    \label{Fig:EffectiveActions}
\end{figure}

To investigate how the Dynamic abstractor works, Figure~\ref{Fig:EffectiveActions} shows the number of effective actions available to the Planner (A*) for each level of abstraction.
We consider an action to be `effective' if its precondition (starting state) is considered different from the outcome at the given abstraction level. At level 0, all actions are different, so the planner has 15,000 actions available at its disposal (all the actions found in the intrinsic phase).
However, it is unlikely that the starting state during the extrinsic phase is exactly the same as any precondition of the action triplets and it is also unlikely that the goal matches the outcome of an action triplet.
So at abstraction level 0, the Planner usually cannot find an action to start from or an action that brings to a a state as the goal state.
As the abstraction level is increased to higher levels, the minimum distance to consider two states as different rises (orange line in the figure), so an increasing number of actions are not effective as the precondition and outcome are no longer considered different.
In particular, we can see that up to about the abstraction level 171 the distance is very small, less than 1 cm.
When the distance is about 1 cm, only 2,435 effective actions remain.
This is because during the intrinsic phase most of the time the random macro-actions produced by the Explorer and performed in the environment miss the object and so the precondition-outcome difference is basically only due to the noise of the perceived object position.
Indeed, only about 1 in 6 times the robot hits the object and displaces it by a distance higher than 1 cm.
In the simulations, the Planner quickly filters out all the first abstraction levels where no solution is found and then starts finding workable solutions at the abstraction level 170 or higher.
The combination of the Planner with the Dynamic Abstraction thus works effectively in providing the right abstraction level with which to plan by adapting to the experienced data and without the need of providing any preset threshold.
In the given domain, this allows the system to automatically adapt to the failure to obtain significant effects with most actions and to the noise of the perceived object position.

\paragraph{REAL-LD: test in the REAL\_IM condition}
The REAL-LD architecture in the REAL\_IM condition achieves an average score of 0.222, even slightly higher than the score of the REAL-D architecture in the simpler REAL\_OM condition (0.216).
This confirms that using the VAE to convert the images into a latent space, and then make plans on such a basis, performs just as well as receiving directly the object positions.

\begin{figure}
    \centering
    \includegraphics[width=8cm]{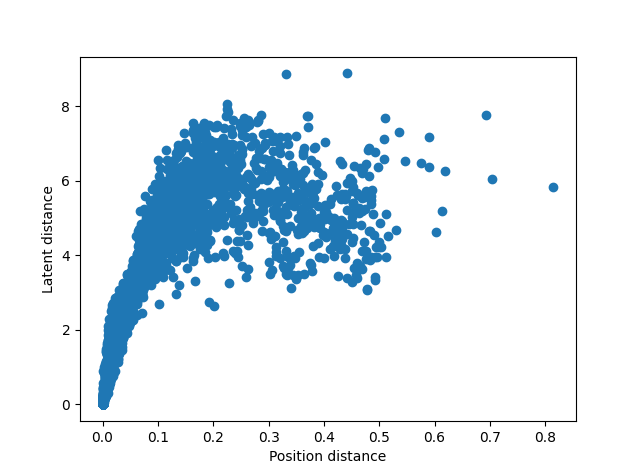}
    \caption{REAL-LD in the REAL\_IM condition: correlation between distances between randomly-chosen couples of object locations (perceived through the RGB images), measured in both the real and the latent space. Spearman R correlation = 0.75, $p < 0.001$; Pearson = 0.83, $p < 0.001$.}
    \label{Fig:LatentDistances}
\end{figure}

\begin{figure}
    \centering
    \includegraphics[width=8cm]{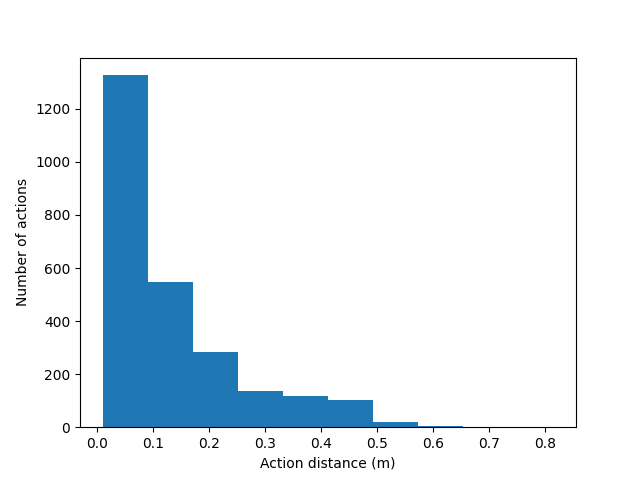}
    \caption{REAL-LD in the REAL\_IM condition: about 80\% of the actions longer than 1 cm are shorter than 20 cm.}
    \label{Fig:HistActions}
\end{figure}

\begin{figure}
    \centering
    \includegraphics[width=8cm]{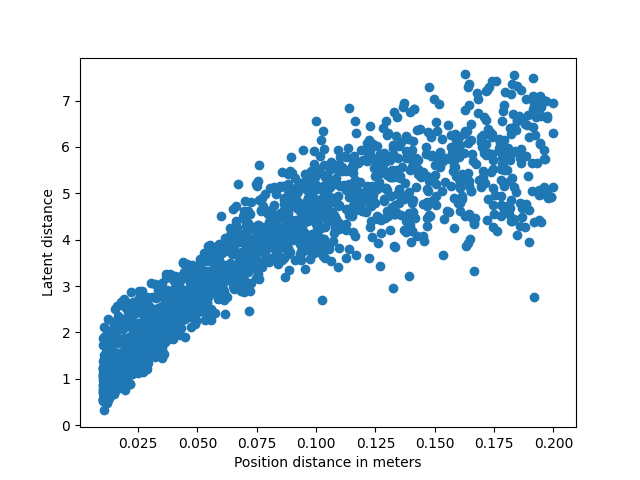}
    \caption{REAL-LD in the REAL\_IM condition: correlation between distances between couples of states, measured in both the real and the latent space, restricted to distances ranging between 1 and 20 cm.
    Spearman R correlation = 0.94, $p < 0.001$; Pearson = 0.90 $p < 0.001$. }
    \label{Fig:LatentDistancesFocus}
\end{figure}

\begin{figure}
    \centering
    \includegraphics[width=9cm]{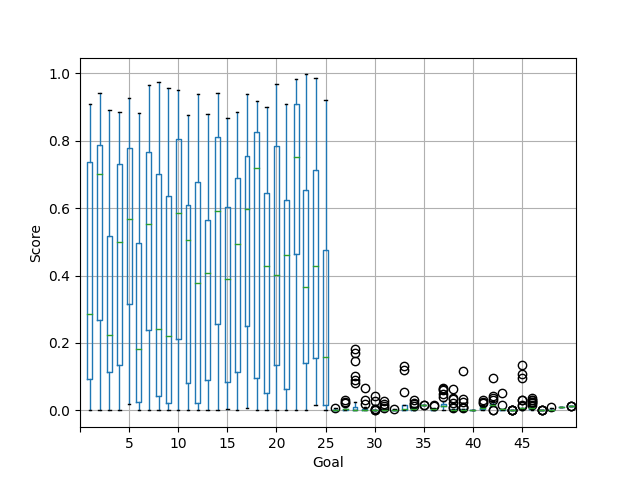}
    \caption{REAL-LD in the REAL\_IM condition: performance per goal. The first 25 goals are 2D goals, while the rest are 2.5D (15) and 3D (10).}
    \label{Fig:REAL-RM-scores}
\end{figure}

We thus investigate the quality of the latent variable representations.
Figure~\ref{Fig:LatentDistances} shows how the Euclidean distance in the latent space correlates with the distance of the object positions.
The correlation is high (Spearman R correlation = 0.75, $p < 0.001$; Pearson correlation = 0.83, $p < 0.001$), especially for short distances.
About 80\% of the actions longer than 1 cm (those mostly used for planning, see preceding analyses) are shorter than 20 cm (see Figure~\ref{Fig:HistActions}): 
Figure~\ref{Fig:LatentDistancesFocus} shows that within that range the correlation is even higher, almost linear (Spearman R correlation = 0.94, $p < 0.001$; Pearson correlation = 0.90 $p < 0.001$).
However, REAL-LD is still limited by the use of the macro-action that can achieve only the first 25 goals but not the other goals involving an object located on the shelf (see Figure~\ref{Fig:REAL-RM-scores}), since the macro-action was bound to push only along the table space.

\begin{figure}
    \centering
    \includegraphics[width=9cm]{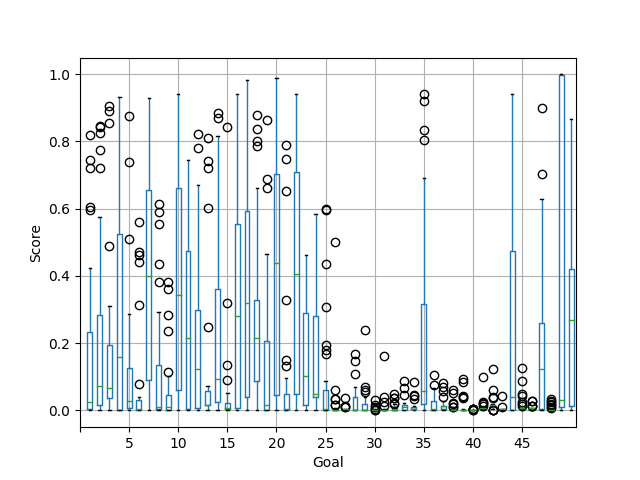}
    \caption{REAL-LD on REAL\_IJ condition: performance per goal. The first 25 goals are 2D goals, while the rest are 2.5D (15) and 3D (10). The system manages to score also on the 2.5D and 3D goals as it can reach objects on the shelf and push them down on the table.}
    \label{Fig:REAL-RJ-scores}
\end{figure}

\paragraph{REAL-LD: test in the REAL\_IJ condition}
REAL-LD was tested in the REAL\_IJ condition to evaluate its ability to handle more general actions defined by an arbitrary sequence of joint via points.
These general joint actions produce non-coordinated arm movements,
which are however still able to hit the object and push it along the table; and they able to reach the shelf as well.
The REAL-LD architecture is capable of extracting from these non-coordinated actions the effective ones to reach its goals, with a performance of 0.139, which is lower than REAL\_OM and REAL\_IM conditions but still higher than the random model (0.038).
Figure~\ref{Fig:REAL-RJ-scores} shows that while the REAL-LD performance is lower with respect to the previous condition with the macro-action, it can still move objects towards their goal.
Moreover, the joint-based action can also reach objects located on the shelf, and this allows REAL-LD to partially accomplish some of the 2.5D and 3D goals (Figure~\ref{Fig:REAL-RJ-scores}).
In particular, the architecture did not find actions to put objects on the shelf but managed to find actions to push them down from the shelf.
These tests thus shows that the REAL-X architecture can be used without predefined macro-actions.
To further improve the performance of REAL-LD in the REAL\_IJ condition, we would need the ability to discover more structured actions (out of the general via-point based actions), but this would probably require a smarter exploration rather than the random-action sampling used by that architecture.

\begin{figure}
    \centering
    \includegraphics[width=8cm]{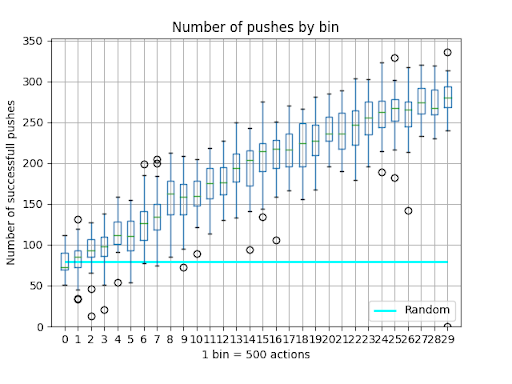}
    \caption{REAL-ILD in the REAL\_IM condition: progressive increase of object contacts/displacements per 500 actions (y-axis), achieved thanks to the IM change predictor used to select actions for exploration during the intrinsic phase (x-axis).}
    \label{fig:real-rmi-intrinsic}
\end{figure}

\paragraph{REAL-ILD: test in the REAL\_IM and REAL\_IJ condition}
Figure~\ref{fig:real-rmi-intrinsic} shows the effectiveness of the change predictor during the intrinsic phase.
Compared to the previous architectures, where random action sampling led to hitting the object only 1 out of 6 actions, REAL-ILD gradually filters out the ineffective actions before executing them, so the chances to actually hit the object rise through the phase up to about 273 hits every 500 actions, so about half of the time.
As an example, one of the simulations had 4269 actions pushing the object for more than 1 cm, compared to the 2435 in the REAL-LD architecture.

\begin{figure}
    \centering
    \includegraphics[width=8cm]{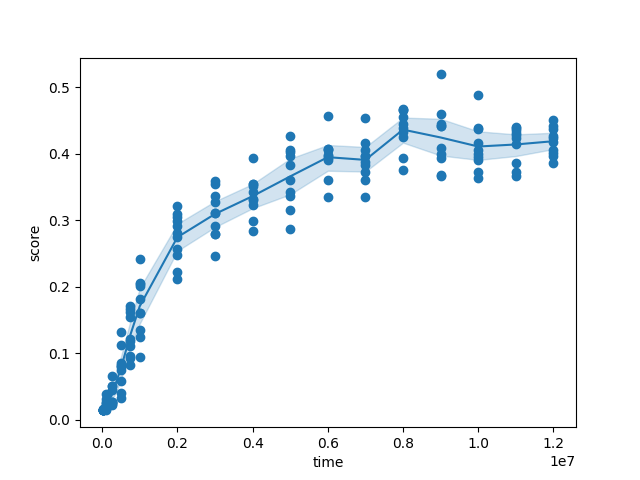}
    \caption{REAL-LD in the REAL\_IM condition: score in the extrinsic phase with different lengths of the intrinsic phase.
    The scores start to plateau after an intrinsic phase of about 8 million time steps, which leads to collect about 8,000 actions.}
    \label{fig:real-score-over-time}
\end{figure}

As shown in Table \ref{Table:SystemsTests},
in the REAL\_IM condition the higher number of actions discovered allows REAL-ILD to achieve a higher performance than REAL-LD, 0.234 vs. 0.222.
In the REAL\_IJ condition, these figures become 0.149 vs. 0.139.
This performance is only marginally higher so we investigated how the REAL-X performance scales with the number of actions.
Figure~\ref{fig:real-score-over-time} shows the performance of REAL-LD on the REAL\_IM condition with different lengths of the intrinsic phase.
We can see that the performance of REAL-LD is not really limited by the number of experienced actions as it is already reaching a plateau with a length of the intrinsic phase of about 8ML steps (8,000 actions).
Most of the performance is indeed acquired during the first 2,000 actions.
These results suggest that to improve the performance the exploration should not simply provide `more actions' but rather focus on finding new actions most needed by the planner. In the case of REAL\_IJ this might for example mean finding actions that have a higher precision and actions able to bring the object to the shelf.

On the other hand, the current approach based on the change predictor can still be very useful where the performance is limited by the small number of actions collected during the intrinsic phase as it improved the speed of finding effective actions up to three-fold.
This can be especially relevant when using real robots, as the exploration can be very time consuming, so any improvement in speed is valuable even if it does not lead to a higher final performance.

\paragraph{REAL-LD with multiple objects}
We also tested REAL-LD using multiple objects. 
With two objects, REAL-LD performance drops to 0.095.
This is still higher than the performance of REAL-R, achieving 0.060, but it is drastically lower than when using only 1 object as the architecture is not equipped to deal with the complexity of multiple objects. 
In particular, multiple objects rise multiple challenges.
First, the combinatorial explosion between the positions of multiple objects worsens the non-reset problem: the robot sees new conditions most of the times, so abstraction with the VAE is more challenging.
Second, moving one object could also move another one, so making the outcomes appear less regular.
We did not investigate which of these causes, or others, represented the most limiting factors.

\section{Related models}
\label{Sec:RelatedModels}

Recently, quite a few articles within the deep-learning literature have proposed systems and algorithms that seem promising to address the challenges posed by the REAL competition. 
We review here these deep-learning systems, many also tested on robotic simulations or on real robots, whose ideas we think could be integrated or compared with those presented above in REAL-X.

Some of these works have focused on exploration.
For example, \cite{Pathak2017, Burda2018} and \cite{Pathak2019} train a policy to explore unknown environments using only images by providing a reward based on prediction errors.
In particular, the policy is rewarded if it leads the agent to places where a predictor gives a high error \cite{Pathak2017, Burda2018}, or where an ensemble of predictors do not agree \cite{Pathak2019} thus showing uncertainty of information.
The exploration reward can be combined with an extrinsic task reward to directly train a desired task policy, or it could be used alone to first explore the environment and then use the data obtained with the exploration to later train a task oriented policy with other methods.
In \cite{Sekar2020} the authors further augment the exploration strategy by adding planning: instead of computing the prediction error retrospectively, they add a planning step where the consequences of the current policy from the current starting state are imagined using a world model, and then the policy is optimised to seek `novel states' before executing it in the environment to explore (similar to our use of a change predictor to prune actions before executing them).
To learn a world model and optimise the policy using imagined trajectories, they rely on the work of \cite{Hafner2019} that shows the advantages of learning long-horizon behaviours based on purely latent variable-based imagination.
In contrast to the previous works, \cite{Sekar2020} uses continuous actions and it learns a world model that is then used to train the agent without additional data for any task for which the reward function is available. The agent uses the world model and the reward function to run a policy in its `imagination' (i.e. running imaginary trials with the world model) and adapt it to the task specified by the reward function.
The trained policy can be then run on the external world without additional training.
However, this requires that the task assigned can be specified with a reward function and that this reward function is provided to the agent.
This is not the case in the REAL benchmark, where only a goal state is provided in the form of an image. An additional component would thus be needed to run the model from \cite{Sekar2020} in REAL, in particular to create a suitable reward function that enables the policy to be trained to reach the target image.

Another work \cite{Pong2019} has also focused on the exploration problem, but in this case the images are first processed through a VAE, and then the exploration is done by drawing from the VAE latent space using biased-weights to encourage exploration of low probability areas.
The same approach of using the latent space of a VAE has been applied earlier in \cite{Nair2018}.
While \cite{Nair2018} did not use biased-weights it showed a full application of the concept on a robotic scenario: starting with a first set of images to train the VAE (either given or obtained by random exploration) the robot automatically samples new images from the VAE and then learns to achieve them, thus setting goals in an autonomous way.
The data collected while trying to achieve those goals can then be further fed back into the VAE, allowing the creation of further new goals.

While the above works base exploration on the `novelty' of states, another work \cite{Florensa2017} instead progressively trains an agent by using a network to predict states of `intermediate difficulty', that is, states that the agent finds possible but hard to reach. This latter work thus grounds its exploration on `competence' instead of `novelty' \cite{SantucciBaldassarreMirolli2013Whichisthebestintrinsicmotivationsignalforlearningmultipleskills}.
However, the system from \cite{Florensa2017} assumes a continuous goal-space representation and so it does not use images but positions and velocities in their state representation. This approach also requires to evaluate the feasibility of each goal by trying to reach multiple times, which can be very time consuming. Furthermore, the model assumes the reset of the environment to its starting condition after every episode, which assumes an external control on the environment to aid the training that is not available in REAL.

Other relevant works have focused on learning suitable representations to enable planning in an unknown environment directly based on images \cite{Ding2020,Laskin2019,Wang2019, Yu2019}.
The model in \cite{Ding2020} uses a mutual information constraint between observations and latent states to train the latent space representations, while at the same time optimising for the rewards by jointly training a reward predictor.
The authors present the results of experiments that show how the learned representations are robust to distractors in the images that do not alter the dynamics of the task. While the focus on mutual information is interesting and could be incorporated in our abstraction module (i.e. in the training objective of the VAE to improve the latent representation), the reward prediction part has no direct application in our intrinsic phase scenario.
The system proposed in \cite{Yu2019} uses a distributional planning network to learn a latent space over which to plan actions to reach a goal.
While the trained distributional planning network they obtain could be directly used as a controller, they then actually train a soft-actor critic controller using rewards derived from the distances in the latent space. Their tests show how the distributional planning network creates a latent space with a distance metric that is more successful and meaningful than the metric learned by an inverse model, a VAE, or by simply calculating the distance in the pixel space. While their simulations and real world experiments use a separate SAC trained for the task, the number of samples used is comparable to our approach (e.g. 20,000 10-frame videos of random interactions; or 28 hours of capture in one of the robotic experiments, which correspond to about 20 million time steps in REAL).

The system proposed in \cite{Laskin2019} is based on a Sparse Graphical Memory (SGM): a new data structure that stores observations and feasible transitions in a sparse memory.
This is similar to our usage of triplets as a graph on which to do planning, but their approach focuses on how to keep the nodes of the graph limited (merging observations) while keeping at the same time the consistency of the graph (e.g. do not merge two states that cannot have similar successors). The system also assume to have access to a short-horizon parametric controller that is capable of accomplishing the task when the starting and goal states are nearby, that is, the optimal action sequence is short.
Similarly to \cite{Wang2019} a low-level controller is thus needed to execute the plans.
The system proposed in this latter work, \cite{Wang2019}, learns a latent space from images with a Causal InfoGAN or a Context Conditional CIGAN to then make a plan with A*. This plan however does not contain actions, but only a sequence of images (reconstructed from the latent variables) of the states that the system has to achieve to reach its goal. This visual plan is then fed to a 'visual tracker', a short-horizon controller that tries to reduce the differences between the current input image and the image of the current step in the visual plan.
Both the inverse model of the visual tracker and the latent space are learned by a dataset that can be constructed by the robot itself with random exploration, with a number of samples comparable to our system (i.e. about 12,000 observations in one of their robotic scenarios). Unfortunately, no code is available to compare their approach to ours.

Mechanisms from all these works might be integrated into the REAL-X architecture components to solve different aspects of the REAL competition, for example to achieve better exploration or to learn better representations.
Among these models, \cite{Laskin2019,Nair2018,Yu2019,Wang2019} seem to offer complete solutions that could be applied to solve the REAL benchmark from the intrinsic phase to the extrinsic phase.
However, there is a relevant technical challenge that is encountered when re-adapting their software to interface it with the benchmark software, which we encountered while trying to apply the code of \cite{Nair2018} to REAL.
The software of this model is indeed organised as commonly done for reinforcement learning systems rather than as it would be required by the REAL open-ended learning benchmark.
In particular, the software requires that the environment variable is passed to the model program which then initialises it and controls it whereas the REAL benchmark requires the model to be passed to the REAL environment (i.e. the REAL evaluation function). The REAL evaluation function also requires that:
(a) the model is organised in terms of an initialisation function and a 1-step function that can be called to pass the last observation to the model and get its next action;
(b) the model can run in an intrinsic motivation mode where the environment is not reset and the model is not given any guidance in terms of reward function, goals, etc.
(c) the model can be run in an extrinsic motivation mode where it can receive goals to pursue within a given time.
When available, the software of most of the models discussed above present this problem and so it is technically cumbersome to adapt their code to test them, or parts of them, with the REAL benchmark software.

This work focused on illustrating and exploring all the several challenges posed by the REAL open-ended learning benchmark, and to develop a general architecture that can hosts different possible solutions to those challenges.
Here we have shown and compare multiple instances of these solutions.
In future work, we plan to use the REAL benchmark to further develop these solutions and systematically compare and integrate them with the models and mechanisms reviewed above.

\section{Conclusions}
\label{Sec:Conclusions}

Several years of research within the developmental robotics community have uncovered important mechanisms for supporting open-ended learning \cite{CangelosiSchlesinger2015DevelopmentalRoboticsfromBabiestoRobots}.
However, this far the community has not managed to propose a benchmark on open-ended learning, allowing quantitative comparison of competing approaches.
The REAL competition closes this gap by formulating an open-ended learning benchmark that allows a rigorous measure of the quality of the knowledge that an autonomous learning agent is able to acquire during free interaction with its environment. 
This measure leverages an extrinsic phase, where an agent is requested to solve a number of tasks sampled from an environment of interest, based on the knowledge that it has autonomously acquired in a previous intrinsic phase involving a long autonomous learning experience in the same environment.
The particular open-ended learning benchmark proposed here involves a camera-arm-gripper robot engaged in manipulating a number of objects on a table and a shelf.
The benchmark is extremely challenging as it requires the solution of a number of problems such as exploring to get in contact with the objects, learning to perceive them, self-generating goals, acquiring the skills to accomplish these goals, etc. 
In addition, the robot has to face all these challenges at the same time, finding itself in a condition similar to the one of newborn infants.

Here we have presented several versions of a robot architecture, REAL-X, that can be used to face different versions of the benchmark where initial simplifications are progressively removed.
The architecture abilities involve the capacities:
to autonomously process images to extract relevant information on objects;
to explore the environment and progressively focus on relevant action outcomes;
to dynamically abstract over state representations in order to support action planning.
The performance of the different versions of the architecture was well above chance level when some simplifications were still kept, and still above chance level in the most challenging conditions. This suggests that the REAL benchmark might remain a challenge for years to come. We are aware that the REAL-X architectures represent only a first step towards a full solution of the benchmark.
Future work might consider longer intrinsic phases to allow the emergence of more sophisticated behaviours, for example for manipulating the objects beyond only pushing them (e.g., for hitting, turning, or grasping them); the introduction of other mechanisms to support a further focusing of the robot's learning processes;
and the introduction of other abstraction or attentional processes allowing the robot to cope with multiple objects.
Notwithstanding these open problems, the REAL benchmark and the REAL-X architectures represent valuable tools to study the fascinating challenges posed by truly open-ended learning.

\section{Acknowledgements}
We thank Francesco Mannella for developing the REAL environment; Simone Asci for developing and testing earlier versions of the change predictor.
We also thank Hewlett Packard Enterprise for providing the hardware to run several of the simulations whose results are reported here.
%This research was supported by the European Union’s Horizon 2020 Research and Innovation Programme under Grant Agreement No 713010, Project “GOAL-Robots – Goal-based Open-ended Autonomous Learning Robots”. 

\bibliographystyle{IEEEtran}
\bibliography{tcds_submission01}

\end{document}